%% file: arxiv_v1.tex
\documentclass[]{style}
\usepackage{graphicx}
\usepackage{pdflscape}
\usepackage[figuresright]{rotating}
\usepackage{tabularray}
\UseTblrLibrary{diagbox}
\usepackage[hidelinks]{hyperref}
\usepackage{marvosym}

\usepackage{comment}
\usepackage{tabularx}
\usepackage{subcaption}
\usepackage{multirow}
\usepackage{xcolor}
\usepackage{amsfonts}
\usepackage{bm}
\usepackage{makecell}
\usepackage{array}

\title{Single-Agent Scaling Fails Multi-Agent Intelligence: Towards Foundation Models with Native Multi-Agent Intelligence }

\author[]{Shuyue Hu$^{*}$}
\author[]{Haoyang Yan$^{*}$}
\author[]{Yiqun Zhang}
\author[]{Yang Chen}
\author[]{Dongzhan Zhou}
\author[]{Lei Bai\textsuperscript{\Letter}}

\affil[]{Shanghai Artificial Intelligence Laboratory\\Shanghai, China}

\affil[]{\{hushuyue, yanhaoyang, zhangyiqun, chenyang4, zhoudongzhan, bailei\}@pjlab.org.cn}

\date{}

\begin{document}
\maketitle

\begingroup
    \long\def\footnotemarkwidth{1em} 

    \renewcommand\thefootnote{\makebox[\footnotemarkwidth][l]{\scalebox{1.5}*}} 
    \footnotetext{Equal contribution.}

    \renewcommand\thefootnote{\makebox[\footnotemarkwidth][l]{\Letter}} 
\endgroup

\vspace{-30pt}

\begin{abstract}
Foundation models (FMs) are increasingly assuming the role of the ``brain'' of AI agents.  While recent efforts have begun to equip FMs with native single-agent abilities---such as GUI interaction or integrated tool use---we argue that the next frontier is endowing FMs with {\em native multi-agent intelligence}. 
We identify four core capabilities of FMs in multi-agent contexts: understanding, planning, efficient communication, and adaptation. Contrary to assumptions about the spontaneous emergence of such abilities, we provide extensive empirical evidence, across 41 large language models and 7 challenging benchmarks, showing that scaling single-agent performance alone does \emph{not} automatically yield robust multi-agent intelligence. To address this gap, we outline key research directions---spanning dataset construction, evaluation, training paradigms, and safety considerations---for building FMs with native multi-agent intelligence.
\end{abstract}

\keywords{Foundation Models, Large Language Models, Multi-Agent Systems}

\section{1  Introduction}\label{sec:intro}

Today, we are at the forefront of a technological revolution powered by  large language models and vision-language models---two major classes of foundation models (FMs). As these models increasingly take on the role of the ``brain'', we are witnessing the rapid rise of FM-based agents. Although the field has \emph{not yet} converged on a precise definition of what constitutes an FM-based agent~\cite{sapkota2025ai,guo2024large}, a clear trend is emerging: moving toward FMs with \emph{native} agent capabilities, rather than relying solely on external scaffolding through agent frameworks or workflows.

For example, OpenAI's Operator~\cite{OpenAI_Operator_2025} is an agent powered by a FM that is trained to interact with graphical user interfaces (e.g., the buttons, menus, and text fields), enabling it to inspect webpages, type, click, and scroll much like human users. Likewise, Kimi’s K2~\cite{team2025kimi} is a FM that integrates tool-use  capabilities into the model itself, enabling autonomous completion of complex tasks (e.g., data analysis or project-level coding) without depending on externally managed agent workflows.

In this paper, we argue for taking the next step: \textbf{building FMs with \textit{native} \textit{multi-agent} intelligence}. As FMs continue to advance and their eco-system expands, it is no longer a question of ``if'' our lives will involve FM-based agents, 
but an \emph{inevitable} matter of ``when'' and ``in what forms.'' 
In such a future, agents and models will no longer operate in isolation; multi-agent environments will become a primary arena where the central scientific and engineering challenges of FM development arise. To meet these realities, FMs must be intrinsically competent at \textit{interacting} with other models, with other agents, and with humans, all beyond single-agent intelligence.

To build models with native multi-agent intelligence, we must first articulate what that intelligence is.  Decades of research in the multi-agent system community offers rich taxonomies and conceptual foundations~\cite{wooldridge2009introduction,dorri2018multi}. Building on this literature, in Section~\ref{sec:mai}, we highlight four core abilities that we believe are essential for FMs in multi-agent contexts: (i) \textbf{understanding}, (ii) \textbf{planning}, (iii) \textbf{efficient communication}, and (iv) \textbf{adaptation}. Crucially, unlike in single-agent settings, each of these abilities requires explicit reasoning about the presence, goals, and behaviors of other entities, as well as the agent’s own interactions with them.

A counter-argument is that explicitly focusing on multi-agent intelligence may be unnecessary; given scaling laws and the emergent properties of FMs, one might expect that strong single-agent abilities would naturally translate into strong multi-agent competence. FM research~\cite{kaplan2020scaling,lee2023rlaif} broadly agrees that performance improves with larger models, more data, and more compute. A related but more contested view holds that once a model surpasses certain scale thresholds,  new capabilities can emerge abruptly~\cite{schaeffer2023emergent}, even when smaller models show no incremental progress toward them. Taken together, these perspectives might suggest that sufficiently advanced FMs will acquire  multi-agent intelligence as a natural \emph{byproduct} of single-agent scaling, without requiring explicit focus.

This counter-argument is theoretically ungrounded~\cite{la2025large}. Multi-agent intelligence is not a trivial extension of single-agent competence; rather, it introduces a set of fundamentally distinct challenges~\cite{kraus1997negotiation,torreno2017cooperative,mu2024multi,haynes2017engineering}, including coordination, bargaining, competition, organization, and norms.
To empirically examine this claim, in Section~\ref{sec:exp},  we consider two open-weight model families (Qwen\footnote{https://huggingface.co/Qwen/models} and LLaMA\footnote{https://huggingface.co/meta-llama/models}) whose development histories are publicly documented. We select 41 models from these two families, spanning parameter scales from 0.5B to 235B and covering releases from 2023 to 2025, and evaluate them on three categories of tasks:
(i) single-agent tasks (\textit{MATH-500}~\cite{lightman2023lets}, \textit{HumanEval}~\cite{chen2021codex}, \textit{MMLU-Pro}~\cite{NEURIPS2024_ad236edc}, and \textit{GPQA}~\cite{rein2024gpqa}) commonly used to measure FM capabilities, (ii) multi-agent understanding tasks (\textit{ToMBench}~\cite{Chen2024ToMBenchBT} and \textit{EmoBench}~\cite{sabour-etal-2024-emobench}) involving theory-of-mind and emotional inference, and (iii) multi-agent planning tasks (\textit{CoordinationQA}~\cite{agashe2025llmcoordinationevaluatinganalyzingmultiagent}) requiring coordination.

Across these 41 models from two families,  we consistently observe that \textbf{scaling single-agent abilities does \emph{not} yield comparable improvements in multi-agent abilities}. More specifically, when comparing newer generations of the same model family at similar parameter sizes (e.g., Qwen-1-1.8B vs. Qwen-3-1.7B), we find that performance on single-agent tasks improves substantially (e.g., accuracy nearly triples from Qwen-1-1.8B to Qwen-3-1.7B), whereas the gains on multi-agent tasks are \emph{far} more modest (e.g., accuracy on the multi-agent planning task increases by only about 30\% from Qwen-1-1.8B to Qwen-3-1.7B).
    

This persistent gap suggests that no matter how large or capable a model becomes in isolation, multi-agent competence requires abilities that single-agent scaling alone does not reliably produce. 
In Section~\ref{sec:disc}, we discuss concrete directions for bridging this divide. We discuss how dataset construction can better capture the rich strategic, social, and organizational dynamics of multi-agent settings; how evaluation protocols must evolve to measure coordination, competition, and emergent group behaviors; and how different training paradigms, ranging from single-model to population-based training, can cultivate the skills needed for robust multi-agent competence. Finally, we highlight safety issues and systemic risks introduced by multi-agent FMs.

\section{2  Multi-Agent Intelligence}\label{sec:mai}

We highlight four core abilities essential for FMs in multi-agent environments:  understanding, planning, efficient communication, and adaptation. While it is impossible to be exhaustive in this paper, we offer them as a preliminary blueprint, and invite the community to refine it.

\paragraph{\textit{Multi-Agent Understanding.}} The ability to understand other entities and their shared environment is fundamental for any meaningful interaction. This requires FMs to be capable of reasoning about others’ beliefs, desires, intentions, and emotions---abilities commonly associated with the theory of mind~\cite{Premack_Woodruff_1978,Kosinski_2024}. Beyond understanding other individuals, multi-agent understanding also entails recognizing shared knowledge embedded in the environment~\cite{savarimuthu2011norm,morris2019norm}, such as social norms, conventions, and communication protocols.

\paragraph{\textit{Multi-Agent Planning.}}
Current FMs demonstrate strong single-agent planning, typically decomposing complex, long-horizon tasks into manageable subtasks. However, multi-agent planning introduces new challenges~\cite{torreno2017cooperative,de2009introduction}---agents  plan while explicitly accounting for the actions, goals, and uncertainties of others. Coordinated planning may require negotiation, alignment of preferences, and anticipation of non-stationary behaviors.
Decentralized multi-agent planning is particularly difficult: agents must generate plans under limited information, asynchronous communication, and rapidly changing contexts. The computational complexity of these problems is often significantly higher than that of single-agent planning, frequently scaling with the number of agents.

\paragraph{\textit{Efficient Communication.}}
Although modern FMs possess fluent natural-language capabilities, efficient communication remains an open challenge. Efficient communication requires conveying information precisely and succinctly~\cite{wang2020learning}. By contrast, state-of-the-art reasoning models frequently produce overly verbose outputs, generating long chains of thought where humans would use less words. Moreover, it is well known that natural language is somewhat redundant and far from maximally efficient~\cite{shannon1951prediction}.  Thus, efficient communication may require more compressed representations and specially designed information-passing mechanisms. Structured protocols—such as learned communication tokens, symbolic messages, or code-like representations—could provide higher bandwidth, reduced ambiguity, and improved coordination efficiency.

\paragraph{\textit{Multi-Agent Adaptation.}} Even agents capable of understanding others, planning jointly, and communicating efficiently, they may still fail to coordinate or compete effectively. This is because multi-agent settings are inherently dynamic: plans may break down due to unexpected behaviors, environmental shifts, or adversarial actions.
Effective multi-agent intelligence therefore requires real-time adaptation: the ability to revise beliefs, update strategies, and adjust interactions on the fly. Adaptive agents continuously integrate new signals, detect deviations from expected behavior, and modify coordination strategies accordingly. This capability is crucial for resilience, robustness, and stable long-horizon interactions.

\begin{figure}
  \centering
\includegraphics[width=0.9\linewidth]{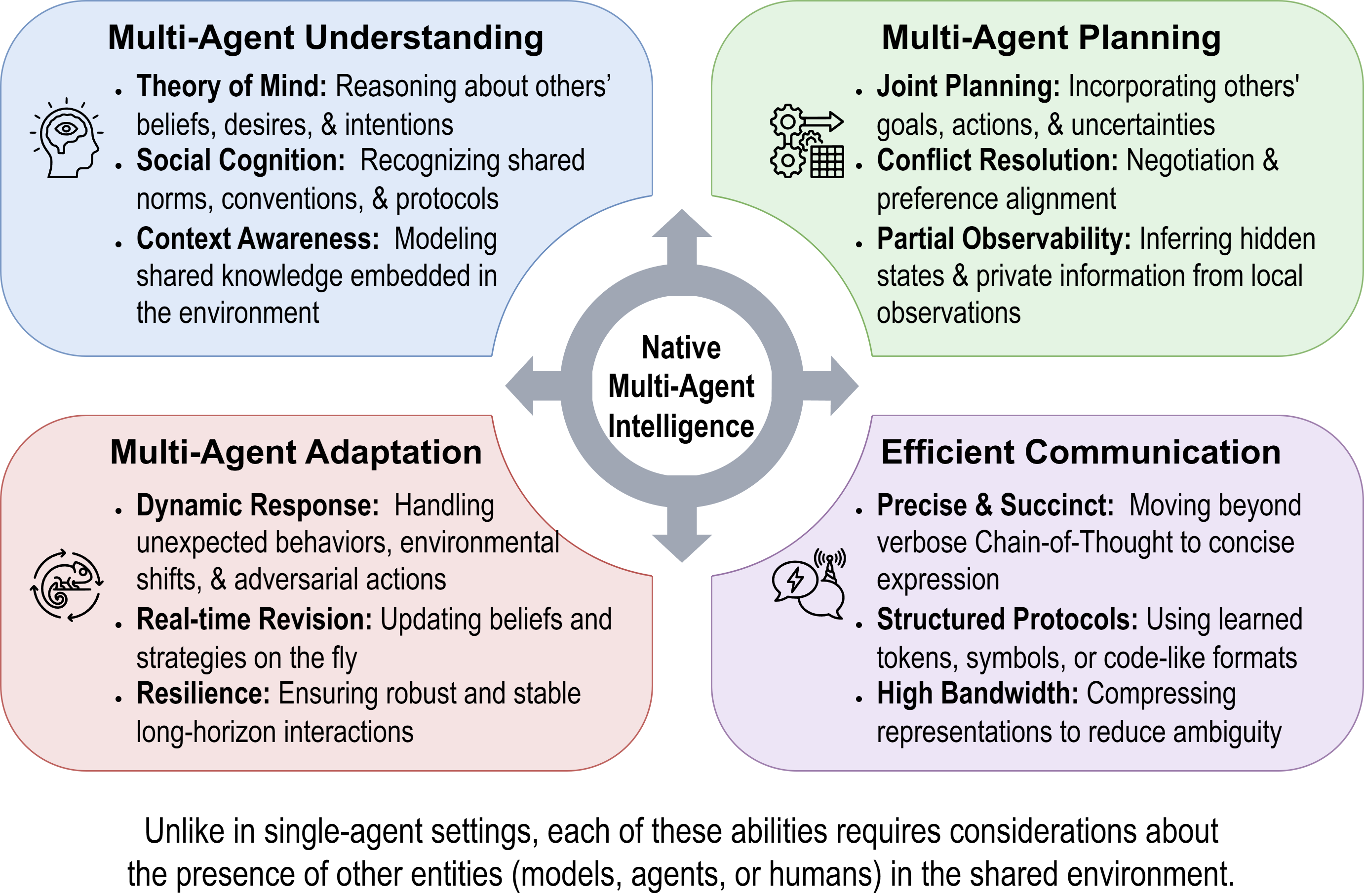}
  \caption{A blueprint for FM with native multi-agent intelligence.}
\end{figure}

\section{3  Experiments}\label{sec:exp}

Our experiments explore the question: do strong single-agent abilities naturally translate into strong multi-agent intelligence? Put differently, we ask if scaling single-agent capabilities alone is sufficient for a model to automatically acquire  multi-agent intelligence.

\subsection{3.1  Experimental Settings}

\paragraph{\textit{Models.}}
We consider two open-weight model families with well-documented development histories: the Qwen family (Qwen, Qwen1.5, Qwen2, Qwen2.5, and Qwen3) and the LLaMA family (LLaMA-2, LLaMA-3, LLaMA-3.1, LLaMA-3.2, and LLaMA-3.3). Together, these account for 41 models in total, spanning parameter scales from 0.5B to 235B and covering releases from 2023 to 2025.
For all models, we utilize the instruction-tuned versions and deploy the official checkpoints via \texttt{vLLM} with the officially recommended hyperparameters.

\paragraph{\textit{Tasks.}}
We evaluate models on 7 challenging benchmarks spanning single-agent (SA) tasks, multi-agent (MA) understanding tasks, and MA planning tasks. 
Specifically, SA benchmarks include: (i) \textit{MATH-500}~\cite{lightman2023lets},  consisting of 500 challenging math problems, (ii) \textit{MMLU-Pro}~\cite{NEURIPS2024_ad236edc}, consisting of 12k questions across various disciplines, such as engineering and business, 
(iii) \textit{HumanEval}~\cite{chen2021codex},  consisting of 164  coding problems, and (iv) \textit{GPQA}~\cite{rein2024gpqa}, consisting of 448 graduate-level questions in biology, physics and chemistry. These benchmarks are widely used to assess FM capabilities and do not require multi-agent abilities. 
For MA understanding tasks, we select: (i) \textit{ToMBench}~\cite{Chen2024ToMBenchBT}, comprising 2,860 questions designed to evaluate theory-of-mind abilities in realistic scenarios, and (ii) \textit{EmoBench}~\cite{sabour-etal-2024-emobench}, consisting of 400 questions targeting emotional understanding. 
For MA planning tasks, we consider \textit{CoordinationQA}~\cite{agashe2025llmcoordinationevaluatinganalyzingmultiagent},  which evaluates multi-agent coordination through 198 questions derived from four coordination games, probing environment comprehension and joint planning.
Performance on all benchmarks is measured using \emph{accuracy} as a unified metric (with \emph{pass@1} used for \textit{HumanEval}). 

\subsection{3.2  Results and Analysis}
\begin{figure*}[t]
  \centering
  \includegraphics[width=0.9\linewidth]{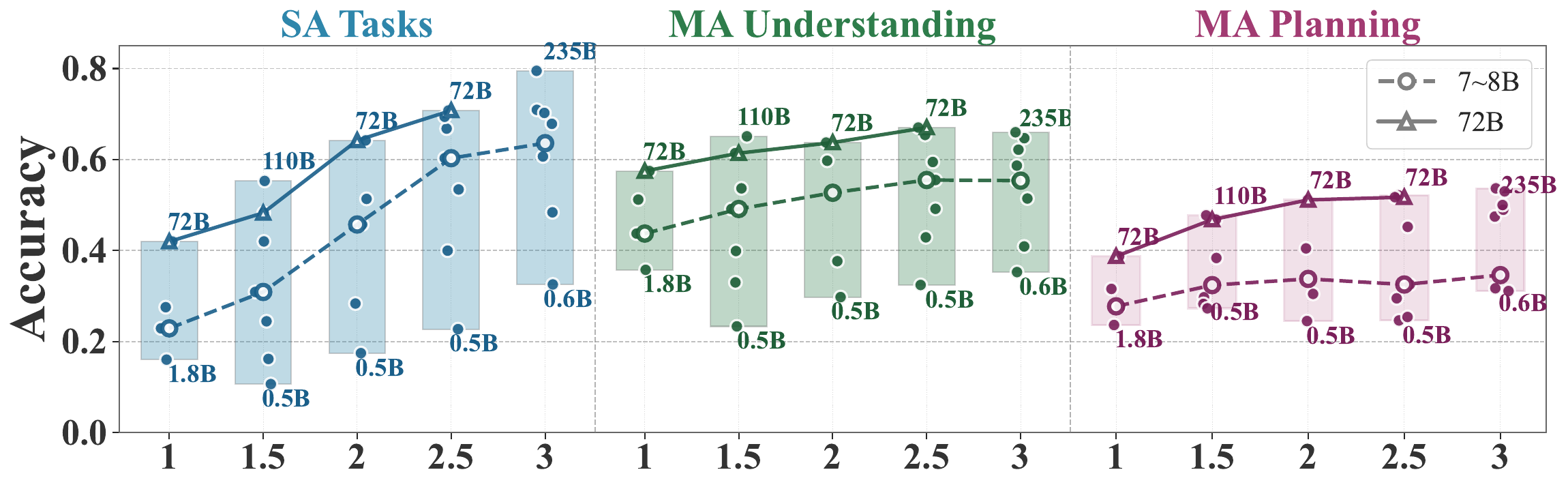}
  \vspace{-5pt}
  \caption*{(a) Qwen Family}

  \includegraphics[width=0.9\linewidth]{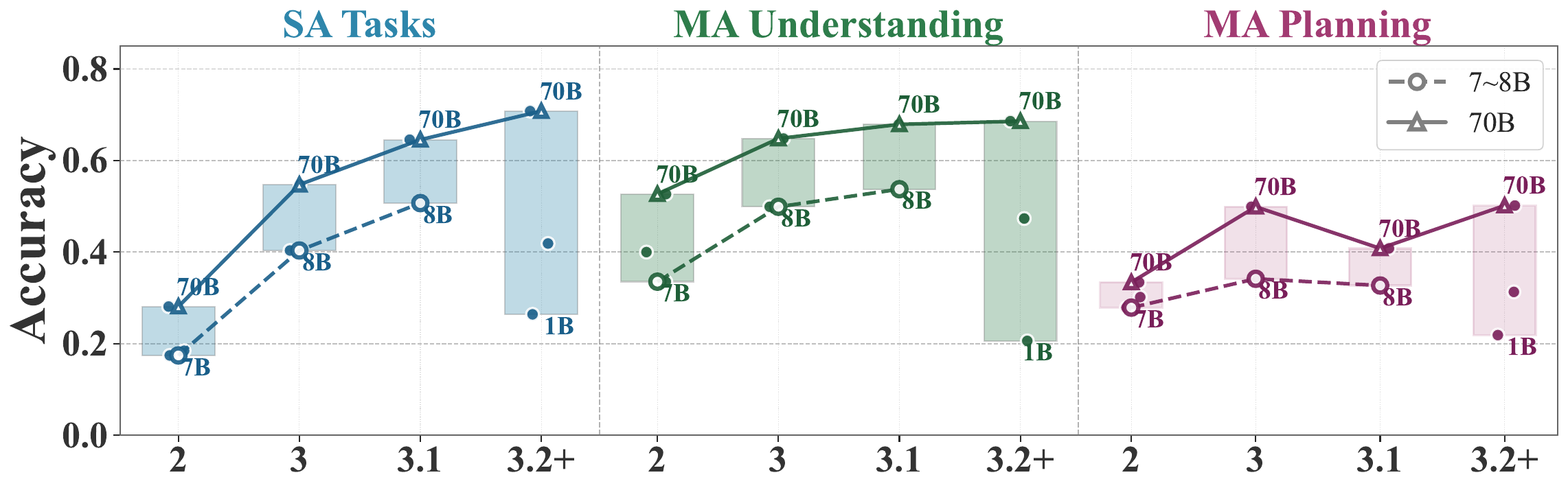}
  \vspace{-5pt}
  \caption*{(b) LLaMA Family}
  \caption{Scaling single-agent abilities does not yield comparable improvements in multi-agent abilities. For both the Qwen and LLaMA families, performance on single-agent (SA) tasks improves substantially across generations for models of similar sizes. In contrast, performance on multi-agent (MA) understanding tasks shows only modest gains, and MA planning performance exhibits little progress.}
  \label{fig:temporal_evolution}
\end{figure*}

Figure~\ref{fig:temporal_evolution} presents the performance of 41 models from the Qwen and LLaMA families in three task categories. Each shaded bar denotes a model generation,
with the shaded region indicating the performance range, and each dot represents the performance of an individual model with a specific parameter size. We highlight the results of models with approximately 8B and 72B parameters to better illustrate performance trends across generations.
For a full listing of all model performances on SA and MA tasks, see Table~1 and Table~2 in Appendix.

Results show that SA tasks exhibit a steep upward trend across generations for models with similar parameter sizes, while improvements on MA understanding tasks remain relatively modest. For example, from Qwen-1 to Qwen-3, the SA task accuracy of approximately 8B models rises sharply from 0.23 to 0.64, while their MA  understanding task accuracy increases only from 0.44 to 0.55. A similar pattern holds for the 72B models: from Qwen-1 to Qwen-2.5, SA task accuracy improves substantially from 0.42 to 0.71, whereas MA understanding task accuracy increases only from 0.57 to 0.67.
Similar trends are also evident within the LLaMA family.

Moreover, for MA planning tasks, performance improvements across generations are even \emph{less} pronounced. In both the Qwen and LLaMA families, the accuracy of $\sim$8B models remains largely flat, stabilizing between 0.2 and 0.35, from Qwen-1 to Qwen-2.5 and from LLaMA-2 to LLaMA-3.1. For $\sim$72B models, accuracy shows little to no upward trend, and in some cases even declines slightly across generations (e.g., from LLaMA-3 to LLaMA-3.1).

\begin{figure}[ht]
  \centering

  \begin{minipage}{0.49\linewidth}
    \centering
    \includegraphics[width=\linewidth]{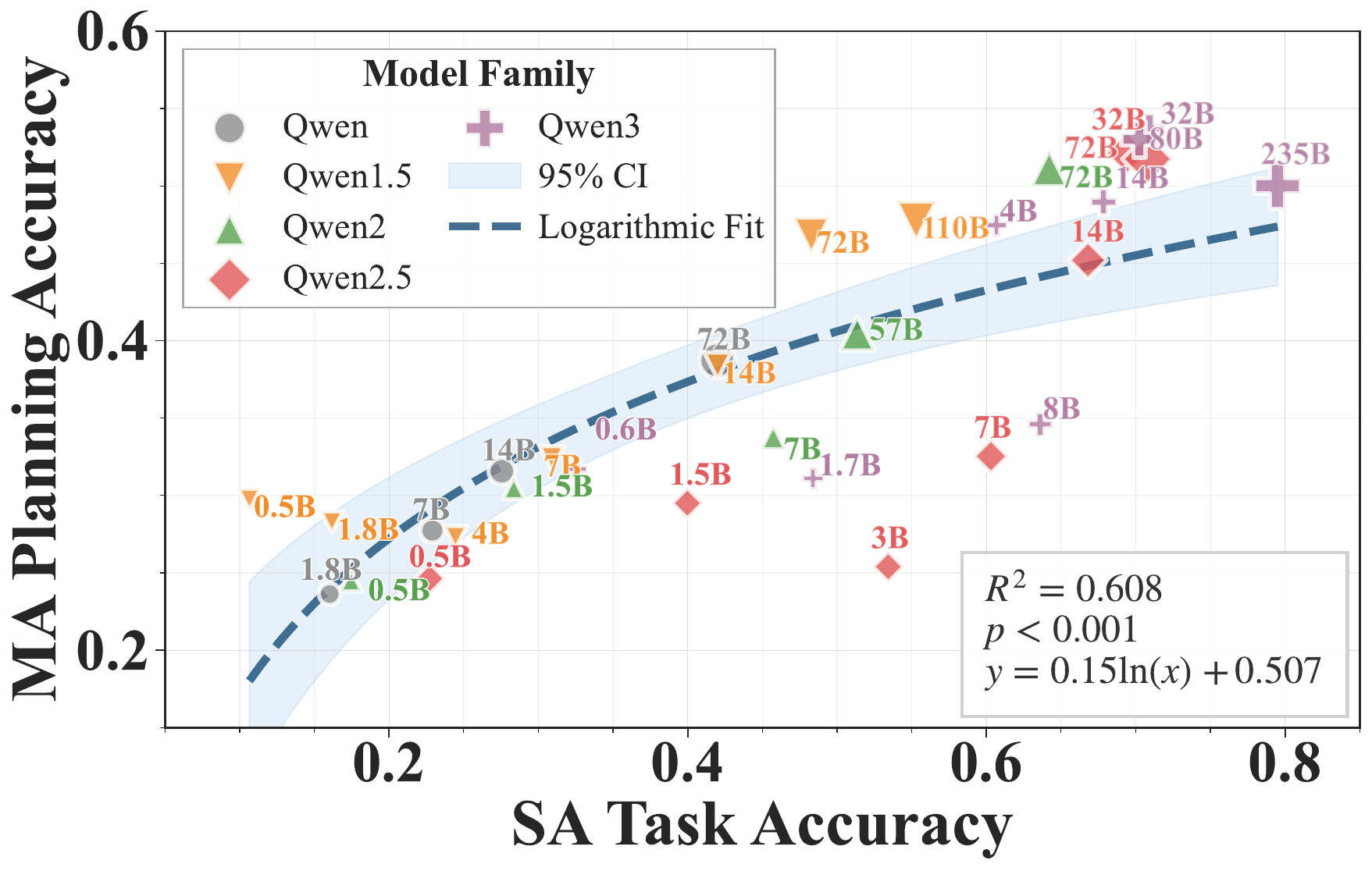}
    \vspace{-5pt}
    \centerline{\small \textbf{(a)} Qwen Family}
  \end{minipage}
  \begin{minipage}{0.49\linewidth}
    \centering
    \includegraphics[width=\linewidth]{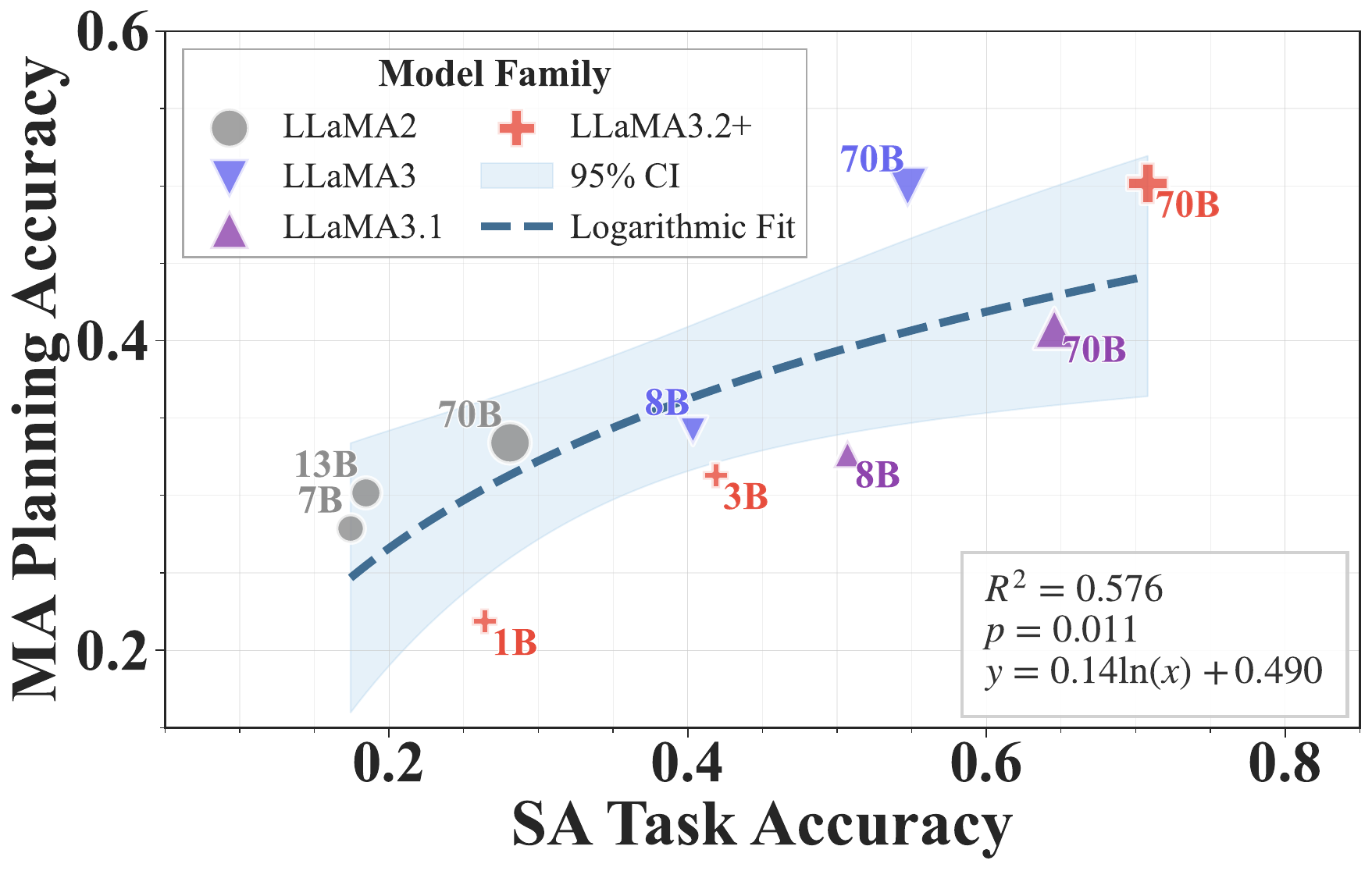}
    \vspace{-5pt}
    \centerline{\small \textbf{(b)} LLaMA Family}
  \end{minipage}


  \caption{Multi-agent intelligence requires capabilities not reliably produced by scaling single-agent tasks. Although SA task accuracy positively correlates with MA planning accuracy, substantial variance remains: models with similar SA task performance often exhibit markedly different MA planning competence. Moreover, the logarithmic form indicates diminishing returns: even substantial increases in SA task accuracy yield only modest improvements in MA planning ability.}
  \label{fig:scaling_correlation}


\end{figure}

Figure~\ref{fig:scaling_correlation} shows how performance on multi-agent planning tasks relates to performance on SA tasks across the two model families. 
Each point represents an individual model, with marker size proportional to its parameter size, and logarithmic regression curves are shown to characterize the overall trend. 
We observe that even models achieving high SA task accuracy (0.6–0.8) exhibit substantial variability in their MA planning task accuracy, suggesting that strong single-agent abilities alone are insufficient for reliably scaling multi-agent planning abilities.

\section{4  What's next?}\label{sec:disc}

\subsection{4.1  Dataset Construction}
Perhaps surprisingly, current FM research lacks datasets that capture multi-agent abilities. Existing resources are highly imbalanced: among the core abilities we outline, multi-agent understanding---especially theory of mind---has received the most attention~\cite{Chen2024ToMBenchBT,gu2024simpletomexposinggapexplicit}, while other abilities remain largely underrepresented.
Interactive environments (e.g., games and simulations) offer a natural source of multi-agent data~\cite{schipper2025pillagerbench,pmlr-v235-li24q}.  However, raw game trajectories are typically ill-suited for large-scale FM training. They can span many steps, include redundant actions, contain extraneous and noisy environmental details, and embed sparse or delayed rewards that provide little instructional signal for FMs. These limitations call for new data-construction strategies tailored to FMs. Promising directions include extracting key decision points from multi-step environments, generating structured interaction episodes that target specific skills (e.g., negotiation, signaling, joint planning), leveraging controlled simulators for systematic synthetic data, and collecting human–human or human–agent interaction data to capture real-world cooperative and competitive behaviors.

\subsection{4.2  Evaluation}
Evaluation faces a similar challenge---current FM research lacks standard benchmarks for evaluating multi-agent capabilities, and among the core abilities we outline, only theory of mind has a relatively mature set of benchmarks. Although various games and simulations exist for evaluating multi-agent planning and adaptation, these environments are not widely adopted~\cite{zhang2024proagent,ma2024large}. One primary reason is that FM research tends to favor Question-Answer~(QA)-style datasets, which allow for fast, single-pass evaluation and straightforward accuracy metrics. In contrast, interactive environments require multi-step execution, substantial token consumption, and repeated trials, making them far less time- and cost-efficient. For example, success rate, one of the most meaningful metrics in these settings, requires multiple repetitions, each costing far more than a single QA.
These limitations highlight the need for benchmark designs that cover a broader range of multi-agent abilities while retaining the efficiency and practicality of QA-style evaluation. At the same time, truly interactive multi-agent testbeds remain essential, yet they demand significant engineering effort for development, validation, and long-term maintenance, which calls for collective community investment.

\subsection{4.3  Training a Single FM} 
Modern FM training pipeline has become relatively standardized and mature~\cite{achiam2023gpt}, yet it remains fundamentally grounded in a single-agent paradigm, and may be insufficient for cultivating native multi-agent intelligence. Multi-agent environments critically hinge on interactions, but defining what constitutes ‘good’ interactive behavior is inherently context-dependent. Unlike tasks with clear ground truth, the quality of an interaction often depends on who the partner is, their goals, preferences, and beliefs, and how these evolve over time. A behavior that is cooperative, aligned, or strategically sound with one partner may be ineffective or even counterproductive with another~\cite{shen2021robust,lu2022model}. This stands in sharp contrast to typical training and alignment paradigms~\cite{sheng2025hybridflow,chu2025sft}, such as supervised finetuning and reinforcement learning from human feedback, all of which require stable, well-specified reward signals that presume a single correct answer or a single preferred trajectory. 
Thus, training models with genuine multi-agent intelligence will require training frameworks that embrace partner-dependent objectives, dynamically evolving feedback, and learning signals derived from diverse inter-agent processes rather than fixed, single-agent labels.

\subsection{4.4  Training a Population of FMs}
Given a fixed compute budget, scaling up a single large model is often more compute-efficient than training many smaller ones. However, in the context of multi-agent environments, training a population of models can become  useful.
First, practical multi-agent deployments---whether for coordination or safety---often require heterogeneous agent roles (e.g., planners, executors, and critics)~\cite{anthropic2025multiagent}. A population enables explicit role specialization while allowing models to learn how to interface with one another. This can yield modular, interpretable multi-agent system architectures that cannot be obtained from a single model.
Second, multi-agent generalization requires diversity~\cite{jaderberg2017population,eccles2019biases}. A single training partner cannot span the vast space of possible beliefs, preferences, or strategies that an agent might face in real-world interactions. A population of models, each differing in capabilities, objectives, or preferences, naturally induces a rich distribution of interaction partners. Training within such a population helps prevent overfitting to a particular partner or reward model, and mitigates brittle behaviors that collapse under distributional shift, thereby promoting generalization and robustness. Third, many multi-agent phenomena are fundamentally population-level and only emerge through rich patterns of inter-agent interactions. Emergent behaviors, such as cooperation and norm formation, arise when individuals interact with multiple others and adapt to a changing strategic landscape~\cite{leibo2017multi}. A population naturally supports such co-evolution: as individuals learn their strategies, they reshape the training environment for others, creating a dynamic feedback loop that drives continual adaptation. 
Taken together,  population-based training of FMs, which is in clear contrast to the dominant paradigm of training a single FM in isolation, might be essential for cultivating multi-agent intelligence.


\subsection{4.5  Safety and Risks}
Safety and risk remain persistent challenges for FMs, and multi-agent environments can introduce new risks that are more complex or qualitatively different than those encountered in single-agent settings. For example, recent studies demonstrate that LLM-based agents may engage in covert collusion through steganographic messaging~\cite{motwani2024secret}, break cooperation promises when playing prisoner's dilemma games~\cite{wang2024large}, and produce or amplify misinformation~\cite{kay2024epistemic}. 
For a detailed taxonomy of risks specific to multi-agent systems, as well as potential research directions for mitigating them, we refer readers to Hammond et al.'s recent work~\cite{hammond2025multi}.

\section{5  Conclusions}\label{sec:conclusion}

Overall, our work adds to a growing body of research recognizing that FM development must embrace the inherently multi-agent nature of future AI systems~\cite{la2025large,cemri2025multi,hammond2025multi,kim2025towards}.
Our key takeaway is that strong multi-agent intelligence will not spontaneously emerge as a byproduct of strong single-agent capabilities, but a frontier in its own right---one that calls for intentional design, sustained innovation, rigorous experimentation, and thoughtful governance.
We invite the community to help refine a shared blueprint for what multi-agent intelligence in FMs should entail, and to explore effective strategies for training models to acquire these abilities.
We believe that the future of AI will be profoundly multi-agent, and that realizing this future will require deliberate, coordinated effort across the field.

\newpage

\bibliographystyle{ref}
\bibliography{reference}

\newpage
\input{appendix}

\end{document}

%% file: appendix.tex


\begin{sidewaystable}[htbp]
\scriptsize
\centering
\renewcommand{\arraystretch}{1.1}
\textbf{\large{Appendix: Model Evaluation Tables}} \\[2mm]
\caption{Performance of Qwen Models on Single-Agent (SA) and Multi-Agent (MA) Tasks}

\begin{tblr}{
  width = \linewidth,
  colspec = {Q[1.1cm,l] Q[3cm,l] Q[1.1cm,c] Q[1.1cm,c] Q[1.1cm,c] Q[1.1cm,c] Q[1.8cm,c] Q[1.8cm,c] Q[1.8cm,c] Q[1.4cm,c] Q[1.4cm,c] Q[1.2cm,c] Q[1.0cm,c]},
  row{1} = {gray9},    
  hlines, vlines,
  cell{1}{1} = {r=2, c=2}{c}, 
  cell{1}{3} = {c=4}{}, 
  cell{1}{7} = {c=3}{}, 
  cell{1}{10} = {c=4}{}, 
  cell{1-Z}{1-Z} = {m},
}

\diagbox[width=5cm, height=5em]{
    \hspace*{1em}\raisebox{1em}{\textbf{\small Models}} 
  }{
    \raisebox{-1em}{\textbf{\small Tasks}}\hspace*{1em}   
  } &                          & \textbf{\small MA Understanding } &                                                         &              &              & \textbf{\small MA Planning } &                     &                    & \textbf{\small SA Task } &          &           &       \\
                                                                &                          & {ToMBench\\(Tasks)}            & {ToMBench\\(Abilities)} & {EmoBench\\(EU)} & {EmoBench\\(EA)} & {CoordinationQA\\(EC)}    & {CoordinationQA\\(ToM)} & {CoordinationQA\\(JP)} & MATH-500          & MMLU-pro & HumanEval & GPQA  \\
\SetCell[r=4]{c} \textbf{Qwen }                    & Qwen-1.8B-Chat                & 0.371                      & 0.403                                                   & 0.165        & 0.505        & 0.339                 & 0.235               & 0.135              & 0.072             & 0.130    & 0.140     & 0.299 \\
                                                                & Qwen-7B-Chat                  & 0.447                      & 0.467                                                   & 0.205        & 0.640        & 0.379                 & 0.318               & 0.135              & 0.120             & 0.222    & 0.299     & 0.275 \\
                                                                & Qwen-14B-Chat                 & 0.518                      & 0.516                                                   & 0.385        & 0.635        & 0.423                 & 0.344               & 0.180              & 0.178             & 0.281    & 0.348     & 0.295 \\
                                                                & Qwen-72B-Chat                 & 0.595                      & 0.609                                                   & 0.445        & 0.670        & 0.529                 & 0.424               & 0.209              & 0.352             & 0.473    & 0.530     & 0.324 \\
\SetCell[r=7]{c}\textbf{Qwen1.5 }                 & Qwen1.5-0.5B-Chat             & 0.281                      & 0.260                                                   & 0.020        & 0.420        & 0.362                 & 0.285               & 0.245              & 0.030             & 0.119    & 0.049     & 0.228 \\
                                                                & Qwen1.5-1.8B-Chat             & 0.350                      & 0.340                                                   & 0.145        & 0.505        & 0.392                 & 0.194               & 0.262              & 0.078             & 0.134    & 0.140     & 0.295 \\
                                                                & Qwen1.5-4B-Chat               & 0.422                      & 0.447                                                   & 0.190        & 0.560        & 0.397                 & 0.264               & 0.158              & 0.108             & 0.226    & 0.305     & 0.339 \\
                                                                & Qwen1.5-7B-Chat               & 0.521                      & 0.544                                                   & 0.305        & 0.625        & 0.424                 & 0.362               & 0.186              & 0.224             & 0.255    & 0.427     & 0.330 \\
                                                                & Qwen1.5-14B-Chat              & 0.581                      & 0.610                                                   & 0.345        & 0.655        & 0.550                 & 0.403               & 0.198              & 0.384             & 0.392    & 0.591     & 0.313 \\
                                                                & Qwen1.5-72B-Chat              & 0.642                      & 0.692                                                   & 0.485        & 0.665        & 0.642                 & 0.447               & 0.315              & 0.440             & 0.511    & 0.652     & 0.328 \\
                                                                & Qwen1.5-110B-Chat             & 0.662                      & 0.733                                                   & 0.485        & 0.735        & 0.665                 & 0.470               & 0.297              & 0.542             & 0.542    & 0.744     & 0.384 \\
\SetCell[r=5]{c}\textbf{Qwen2 }                   & Qwen2-0.5B-Instruct               & 0.328                      & 0.348                                                   & 0.090        & 0.455        & 0.376                 & 0.245               & 0.114              & 0.110             & 0.096    & 0.195     & 0.297 \\
                                                                & Qwen2-1.5B-Instruct               & 0.420                      & 0.420                                                   & 0.150        & 0.560        & 0.470                 & 0.314               & 0.130              & 0.216             & 0.203    & 0.427     & 0.288 \\
                                                                & Qwen2-7B-Instruct                 & 0.576                      & 0.581                                                   & 0.345        & 0.655        & 0.483                 & 0.371               & 0.159              & 0.532             & 0.260    & 0.720     & 0.317 \\
                                                                & Qwen2-72B-Instruct                & 0.677                      & 0.692                                                   & 0.500        & 0.720        & 0.652                 & 0.524               & 0.358              & 0.708             & 0.650    & 0.817     & 0.393 \\
                                                                & Qwen2-57B-A14B-Instruct           & 0.601                      & 0.622                                                   & 0.460        & 0.710        & 0.548                 & 0.464               & 0.202              & 0.492             & 0.529    & 0.689     & 0.344 \\
\SetCell[r=7]{c}\textbf{Qwen2.5 }                 & Qwen2.5-0.5B-Instruct             & 0.347                      & 0.323                                                   & 0.155        & 0.495        & 0.383                 & 0.227               & 0.129              & 0.290             & 0.117    & 0.268     & 0.234 \\
                                                                & Qwen2.5-1.5B-Instruct             & 0.443                      & 0.462                                                   & 0.235        & 0.590        & 0.455                 & 0.291               & 0.139              & 0.508             & 0.289    & 0.518     & 0.284 \\
                                                                & Qwen2.5-3B-Instruct               & 0.541                      & 0.570                                                   & 0.280        & 0.625        & 0.400                 & 0.268               & 0.094              & 0.646             & 0.452    & 0.744     & 0.295 \\
                                                                & Qwen2.5-7B-Instruct               & 0.580                      & 0.655                                                   & 0.355        & 0.655        & 0.470                 & 0.329               & 0.177              & 0.744             & 0.574    & 0.768     & 0.326 \\
                                                                & Qwen2.5-14B-Instruct              & 0.626                      & 0.659                                                   & 0.440        & 0.685        & 0.535                 & 0.527               & 0.294              & 0.810             & 0.646    & 0.848     & 0.368 \\
                                                                & Qwen2.5-32B-Instruct              & 0.706                      & 0.722                                                   & 0.510        & 0.730        & 0.658                 & 0.573               & 0.333              & 0.830             & 0.698    & 0.866     & 0.382 \\
                                                                & Qwen2.5-72B-Instruct              & 0.695                      & 0.746                                                   & 0.525        & 0.740        & 0.708                 & 0.497               & 0.347              & 0.838             & 0.721    & 0.872     & 0.400 \\
\SetCell[r=8]{c}\textbf{Qwen3 }                   & Qwen3-0.6B               & 0.409                      & 0.418                                                   & 0.175        & 0.465        & 0.421                 & 0.377               & 0.153              & 0.550             & 0.251    & 0.220     & 0.281 \\
                                                                & Qwen3-1.7B               & 0.492                      & 0.527                                                   & 0.180        & 0.520        & 0.450                 & 0.306               & 0.177              & 0.728             & 0.428    & 0.494     & 0.286 \\
                                                                & Qwen3-4B                 & 0.601                      & 0.623                                                   & 0.280        & 0.640        & 0.730                 & 0.430               & 0.264              & 0.792             & 0.595    & 0.701     & 0.339 \\
                                                                & Qwen3-8B                 & 0.653                      & 0.681                                                   & 0.315        & 0.665        & 0.574                 & 0.273               & 0.191              & 0.806             & 0.650    & 0.726     & 0.362 \\
                                                                & Qwen3-14B                & 0.669                      & 0.705                                                   & 0.385        & 0.670        & 0.718                 & 0.453               & 0.298              & 0.842             & 0.687    & 0.774     & 0.411 \\
                                                                & Qwen3-32B                & 0.688                      & 0.720                                                   & 0.445        & 0.700        & 0.823                 & 0.444               & 0.343              & 0.856             & 0.725    & 0.841     & 0.415 \\
                                                                &  Qwen3-235B-A22B-Instruct-2507 & 0.722                      & 0.751                                                   & 0.485        & 0.705        & 0.682                 & 0.485               & 0.333              & 0.920             & 0.830    & 0.957     & 0.473 \\
                                                                & Qwen3-Next-80B-A3B-Instruct       & 0.718                      & 0.745                                                   & 0.520        & 0.715        & 0.742                 & 0.525               & 0.323              & 0.952             & 0.812    & 0.622     & 0.424 
\end{tblr}
\label{tab:qwen_table}
\end{sidewaystable}

\begin{sidewaystable}[htbp]
\scriptsize
\centering
\renewcommand{\arraystretch}{1.1}
\caption{Performance of LLaMA Models on Single-Agent (SA) and Multi-Agent (MA) Tasks}

\begin{tblr}{
  width = \linewidth,
  colspec = {Q[1.2cm,l] Q[3.2cm,l] Q[1.1cm,c] Q[1.1cm,c] Q[1.1cm,c] Q[1.1cm,c] Q[1.8cm,c] Q[1.8cm,c] Q[1.8cm,c] Q[1.4cm,c] Q[1.4cm,c] Q[1.2cm,c] Q[1.0cm,c]},
  row{1} = {gray9},    
  hlines, vlines,
  cell{1}{1} = {r=2, c=2}{c}, 
  cell{1}{3} = {c=4}{}, 
  cell{1}{7} = {c=3}{}, 
  cell{1}{10} = {c=4}{}, 
  cell{1-Z}{1-Z} = {m},
}

\diagbox[width=5cm, height=5em]{
    \hspace*{1em}\raisebox{1em}{\textbf{\small Models}} 
  }{
    \raisebox{-1em}{\textbf{\small Tasks}}\hspace*{1em}   
  } &                          & \textbf{\small MA Understanding } &                                                         &              &              & \textbf{\small MA Planning } &                     &                    & \textbf{\small SA Task } &          &           &       \\
                                                                &                          & {ToMBench\\(Tasks)}            & {ToMBench\\(Abilities)} & {EmoBench\\(EU)} & {EmoBench\\(EA)} & {CoordinationQA\\(EC)}    & {CoordinationQA\\(ToM)} & {CoordinationQA\\(JP)} & MATH-500          & MMLU-pro & HumanEval & GPQA  \\
\SetCell[r=3]{c}\textbf{LLaMA2}   & Llama-2-7b-chat-hf                                    & 0.339 & 0.326 & 0.145 & 0.535                                         & 0.403    & 0.295     & 0.138                                   & 0.068    & 0.178    & 0.152     & 0.299                       \\
                                                    & Llama-2-13b-chat-hf                                   & 0.459 & 0.465 & 0.155 & 0.580                                         & 0.452    & 0.333     & 0.120                                   & 0.088    & 0.210    & 0.159     & 0.281                       \\
                                                    & Llama-2-70b-chat-hf                                   & 0.538 & 0.595 & 0.325 & 0.660                                         & 0.418    & 0.398     & 0.186                                   & 0.142    & 0.316    & 0.360     & 0.306                       \\ 
\cline{1-1}
\SetCell[r=2]{c}\textbf{LLaMA3}   & Meta-Llama-3-8B-Instruct                                    & 0.543 & 0.572 & 0.270 & 0.655                                         & 0.512    & 0.377     & 0.136                                   & 0.302    & 0.415    & 0.598     & 0.299                       \\
                                                    & Meta-Llama-3-70B-Instruct                                   & 0.669 & 0.694 & 0.515 & 0.735                                         & 0.692    & 0.497     & 0.308                                   & 0.492    & 0.621    & 0.701     & 0.375                       \\
\SetCell[r=2]{c}\textbf{LLaMA3.1} & Llama-3.1-8B-Instruct                                  & 0.600 & 0.622 & 0.330 & 0.660                                         & 0.439    & 0.365     & 0.177                                   & 0.518    & 0.501    & 0.665     & 0.344                       \\
                                                    & Llama-3.1-70B-Instruct                                 & 0.716 & 0.737 & 0.565 & 0.735                                         & 0.526    & 0.427     & 0.271                                   & 0.680    & 0.693    & 0.805     & 0.404                       \\
\SetCell[r=2]{c}\textbf{LLaMA3.2} & LlaMA-3.2-1B-Instruct                                  & 0.373 & 0.378 & 0.010 & 0.230                                         & 0.320    & 0.268     & 0.068                                   & 0.300    & 0.190    & 0.311     & 0.255                       \\
                                                    & LlaMA-3.2-3B-Instruct                                  & 0.521 & 0.545 & 0.235 & 0.640                                         & 0.421    & 0.345     & 0.173                                   & 0.494    & 0.386    & 0.500     & 0.295                       \\ 
\cline{1-1}
\SetCell[r=1]{c}\textbf{LLaMA3.3}                                            & LlaMA-3.3-70B-Instruct                                 & 0.726 & 0.738 & 0.560 & 0.760                                         & 0.658    & 0.492     & 0.355                                   & 0.750    & 0.718    & 0.848     & 0.516                      

\end{tblr}{}
\label{tab:qwen_table}
\end{sidewaystable}